\documentclass[11pt]{article}
\usepackage{geometry}
\usepackage{coling2020}
\usepackage{times}
\usepackage{url}
\usepackage{latexsym}
\usepackage{microtype}
\usepackage[export]{adjustbox}
\usepackage{graphicx}
\usepackage{fancyhdr,amsmath,amssymb}
\usepackage[ruled,vlined]{algorithm2e}
\include{pythonlisting}
\usepackage{subcaption}
\usepackage[english]{babel}
\usepackage[utf8]{inputenc}
\usepackage[noend]{algpseudocode}
\usepackage{epstopdf}
\usepackage{appendix}
\usepackage{hyperref}
\usepackage{float}
\usepackage{dirtytalk}
\colingfinalcopy 

\title{Voice@SRIB at SemEval-2020 Task 9 and 12: Stacked Ensembling method for Sentiment and Offensiveness detection in Social Media}

\author{Abhishek Singh \\
  Samsung R\&D Bangalore \\
  {\tt abhi3.singh@samsung.com} \\\And
  Surya Pratap Singh Parmar \\
  Samsung R\&D Bangalore \\
  {\tt s.singhparm@samsung.com} \\}

\begin{document}
\maketitle
\begin{abstract}
 In social-media platforms such as Twitter, Facebook, and Reddit, people prefer to use code-mixed language such as Spanish-English, Hindi-English to express their opinions. In this paper, we describe different models we used, using the external dataset to train embeddings, ensembling methods for Sentimix, and OffensEval tasks. The use of pre-trained embeddings usually helps in multiple tasks such as sentence classification, and machine translation. In this experiment, we have used our trained code-mixed embeddings and twitter pre-trained embeddings to SemEval tasks. We evaluate our models on macro F1-score, precision, accuracy, and recall on the datasets. We intend to show that hyper-parameter tuning and data pre-processing steps help a lot in improving the scores. In our experiments, we are able to achieve 0.886 F1-Macro on OffenEval Greek language subtask post-evaluation, whereas the highest is 0.852 during the Evaluation Period. We stood third in Spanglish competition with our best F1-score of 0.756. Codalab username is asking28.
\end{abstract}

\section{Introduction}
\label{intro}

%
%
\blfootnote{
    %
    %
    %
    %
    %
    %
    \hspace{-0.65cm}  
    This work is licensed under a Creative Commons 
    Attribution 4.0 International License.
    License details:
    \url{http://creativecommons.org/licenses/by/4.0/}.
}

SemEval Task-9 Sentimix \cite{patwa2020sentimix} is divided into two tasks, one for Hinglish (Hindi-English) and the other for Spanglish (Spanish-English) code-mixed subtasks. In the Spanglish task, the dataset contains tweets in Spanglish (Spanish-English) code-mixed language, and it is labeled into three categories positive, negative, and neutral sentiments. The task is to classify codemixed tweets into these three sentiments. SemEval Task-12 OffensEval \cite{zampieri-etal-2020-semeval} is divided into different subtasks, English \cite{rosenthal2020}, Danish \cite{sigurbergsson2020offensive}, Arabic \cite{mubarak2020arabic}, Turkish \cite{coltekikin2020}, and Greek \cite{pitenis2020} languages. English task is divided into three subtasks- A, B, and C.  Subtask-A of OffensEval is offensive language identification, subtask-B is categorization of offensive types into targeted and untargeted, and subtask-C is offensive target identification as individual, group, or other.\par
In the last decade, there has been proliferation in the use of social media web sites. It has led to pervasive use of hate inducing speech and offensive language to express opinions. The use of profane language has been growing in face-to-face interactions as well as online communications in recent years. The anonymity provided by these websites and lack of stringent action has led to adoption aggressive behavior by people.Youth who experienced cyberbullying, as either an offender or a victim, had more suicidal thoughts and were more likely to attempt suicide than those who had not experienced such forms of peer aggression \cite{article}. Hence it's necessary to auto-remove offensive and profane language in an online environment. \par
Since the inflow of such type of content is huge, manual filtering is time-consuming and requires much manual labor; hence it becomes almost impractical to do manual filtering. Due to this reason, researchers have proposed methods to automate filtering process by training machine learning models in pre-annotated datasets hate speech and offensive language by \cite{hateoffensive} \cite{DBLP:journals/corr/abs-1712-06427} , cyberbullying \cite{Xu:2012:LBT:2382029.2382139} and detection of racism by \cite{DBLP:journals/corr/TulkensHLVD16}.\par
In this work (team name is SRIB2020), we try to classify twitter tweets in different languages code-mixed for Sentimix tasks and Monolingual tweets in different languages in the OffensEval task into different classes. In OffensEval tasks, tweets are classified as offensive or non-offensive, whereas in the Sentimix task, tweets are classified as positive, negative, and neutral. In the Sentimix task, the \say{neutral} class is bit ambiguous as many positive tweets in the dataset are labeled as neutral, and negative tweets are labeled as neutral. Class \say{neutral} has a very thin boundary with the other two classes, \say{positive} and \say{negative}.\par
\begin{enumerate}
    \item ID-7229 - \textit{WOO hoo Cricket world cup starts today. Good luck to host @englandcricket hope for a good start.} - This sentence is positive in tone, but it is labeled as neutral.
    \item ID-8199- \textit{@hardikpandya7 best wishes for WorldCup and Eid-Mubarak from MUJAFFAR  Hasan National General Secretary LJP URL}- This tweet is positive in its sentiment but is labeled as neutral.
\end{enumerate}
\cite{lal2019mixing} first generates subword level representations for the sentences using a CNN architecture. The generated representations are used as inputs to a Dual Encoder Network, consisting of two different BiLSTMs - the Collective and Specific Encoder. The Collective Encoder captures the overall sentiment of the sentence, while the Specific Encoder utilizes an attention mechanism to focus on individual sentiment-bearing sub-words. \cite{sharma2016shallow}have annotated the data, developed a language identifier, a normalizer, a part-of-speech tagger, and a shallow parser for sentiment analysis of code-mixed data. \cite{pravalika2017domain} used a lexicon lookup approach to perform domain-specific sentiment analysis. \cite{joshi2016towards}  introduce learning sub-word level representations in LSTM (Subword-LSTM) architecture instead of character-level or word-level representations; this enables to learn the information about sentiment value of meaningful morphemes. \cite{choudhary2018sentiment} uses the shared parameters of siamese networks to map the sentences of code-mixed and standard languages to a common sentiment space. They introduce a primary clustering-based preprocessing method to capture variations of code-mixed transliterated words.
\par
Supervised learning techniques for hate detection, offensive detection, and target and sentiment classification on social media datasets have been explored in recent times. \cite{davidson2017automated} described a way of multi-class classification of offensive language and hate speech in tweets, using SVM, random forest, naive Bayes, and Logistic Regression. \cite{del2017hate} reported performance for a simple LSTM classifier not better than an ordinary SVM, when evaluated on a small sample of Facebook data for only two classes (Hate, No-Hate), and three different levels of strength of hatred. \cite{pitsilis2018detecting} propose a detection scheme that is an ensemble of Recurrent Neural Network (RNN) classifiers. It incorporates various features associated with user-related information, such as the users’ tendency towards racism or sexism.\par
This paper can be summarised into five key points-\par
\begin{enumerate}
\item Applied a variation of Focal Loss by applying class weight along with Gamma parameter in the loss function.
    \item Applied multiple preprocessing on the raw text, since in social media platforms people tend to use incorrect grammatical forms and incorrect spellings, it helped us to increase F1 scores.
    \item For Hinglish subtask we trained our own word embedding by collecting code-mixed datasets from multiple sources.
    \item Ensemble model made of multiple deep-learning based models, CNN, LSTM, and Sequential self-attention on LSTM.
    \item Comparing model performance on different Machine Learning and Deep Learning models.
\end{enumerate}
Rest of the paper is organized as follows:  \textbf{Section-\ref{methodology}} presents the methodology in our paper, data description, pre-processing steps, model description, and parameter tuning. \textbf{Section-\ref{experiments}} presents various experiments performed on different models and their results. Finally in \textbf{Section-\ref{conclusion}} conclusion based on experiments performed and the future work is discussed. Code is available at github\footnote{\url{https://github.com/asking28/sentimix2020}}. 
\section{Methodology}\label{methodology}
\subsection{Data Description}
\begin{enumerate}
    \item Sentimix Tasks- 
    \begin{itemize}
        \item Hinglish test set 3000 unlabeled tweets.
        \item Spanglish test data contains 3789 unlabelled tweets.


\begin{table}[h]
\centering
\begin{tabular}{|c|c|c|c|c|}
\hline
 &
  \textbf{\begin{tabular}[c]{@{}c@{}}Positive\\ Train/Dev\end{tabular}} &
  \textbf{\begin{tabular}[c]{@{}c@{}}Negative\\ Train/Dev\end{tabular}} &
  \textbf{\begin{tabular}[c]{@{}c@{}}Neutral\\ Train/Dev\end{tabular}} &
  \textbf{\begin{tabular}[c]{@{}c@{}}Total\\ Train/Dev\end{tabular}} \\ \hline
\textbf{Hinglish} &
  4634/982 &
  4102/1128 &
  5264/890 &
  14000/3000 \\ \hline
\textbf{Spanglish} &
  6005/1498 &
  2023/506 &
  3974/994 &
  12002/2998 \\ \hline
\end{tabular}
\caption{Training and Dev data distribution Sentimix}
\label{tab:Sentimix data Sitribution}
\end{table}
        
    \end{itemize}
    \item Offenseval Tasks- \par 
    Offenseval English task is divided into three subtasks A, B, and C.
\begin{table}[ht]
\centering
\begin{tabular}{|l|l|l|}
\hline
\textbf{}               & \textbf{Train} & \textbf{Test} \\ \hline
\textbf{Eng. SubTask-A} & 8696199        & 3877          \\ \hline
\textbf{Eng. SubTask-B} & 188974         & 1722          \\ \hline
\textbf{Eng. Subtask-C} & 188974         & 1722          \\ \hline
\textbf{Danish}         & 2961           & 329           \\ \hline
\textbf{Turkish}        & 31756          & 3528          \\ \hline
\textbf{Greek}          & 8743           & 1544          \\ \hline
\textbf{Arabic}         & 7000           & 2000          \\ \hline
\end{tabular}
\caption{OffensEval data distribution}
\label{tab:offenseval_data}
\end{table}
\end{enumerate}

\subsection{Data Preprocessing}
\textbf{Hinglish Data Processing}
\begin{itemize}
     \item \textbf{In Demojisation step}, different types of emojis present in the corpus is converted into corresponding text representation. Since these combined datasets contain large number of tweets and contain different types of emojis, it becomes necessary to convert emojis into corresponding text representations using the cheatsheet list \footnote{\url{https://www.webfx.com/tools/emoji-cheat-sheet/}}. 
     \item \textbf{Removing different types of patterns} such as URLs were replaced with URL  token in the dataset, @USERNAME was converted to USER token and hashtags, \# symbol was removed from the dataset. The dataset is cleaned for different punctuation marks, as punctuation marks are not needed to train the embeddings.
     \item \textbf{Acronyms and Contractions} were replaced with their corresponding English words. We replace it by creating a dictionary of acronyms and contractions mapping to their expanded form. Acronyms such as 4ever are converted to forever, abt to about, cb to comeback, etc. These acronyms are commonly used in social media platforms. Contractions such as can't, aren't, i've, etc. were again converted to their corresponding text cannot, are not, and I have for this case. \par
     
 \end{itemize}
\textbf{Spanglish Data Processing}- The NLTK Snowball Stemmer \footnote{\url{https://www.nltk.org/_modules/nltk/stem/snowball.html}} package was used because it offers to stem in both English and Spanish. The flexibility to use the stemmer in both languages played a key role in the Spanglish Sentiment Analysis system. The list of stop words was constructed from the stop words corpus provided in NLTK. While pre-processing     tokenized tweets, any word included in the NLTK English stop word corpus is excluded. Close attention is paid to elongated words (i.e. – \say{helloooooo} ,  \say{orrrrrale}), and after considering possible features of elongated words, spelling normalization is applied to these tokens.  It would also be beneficial to apply spelling normalization to slang or purposely misspelled words that are common in tweets or other informally written texts. We removed character repetition by removing characters that occurred more than two times continuously. Emoticons are replaced with their corresponding text in the tweets.

\textbf{English Data Preprocessing}- Pre-processing steps such as emoticon replacement, contraction replacement, acronym replacements are done in a similar manner as in previous datasets. In social media platforms, people tend to use short forms such as \textbf{forget} maybe written as \textbf{frgt}. So to deal with this problem we have applied multiple spell correction steps. We have used PySpellchecker \footnote{\url{https://github.com/barrust/pyspellchecker}}, it uses a Levenshtein Distance algorithm to find permutations within an edit distance of 2 from the original word. Then it compares all permutations (insertions, deletions, replacements, and transpositions) to known words in a word frequency list. Those words that are found more often in the frequency list are more likely the correct results. Then we delete characters having more than two continuous occurrences, as it is very rare that a character occurs more than twice continuously.\par
\textbf{Turkish Data preprocessing}- We have followed pre-processing steps as mentioned above with an extra step of Turkish word lemmatization using lemmatization model by \cite{sak2008turkish} which is trained by nearly one million Turkish sentences.\par
\textbf{Arabic, Danish, and Greek Data processing}- Arabic data is first transliterated to Roman script using Classic Language Toolkit (CLTK)\footnote{\url{http://docs.cltk.org/en/latest/}} and then all the steps used for other languages are applied to datasets. Danish data was used as it is. We applied Greek Stemmer from \footnote{\url{https://deixto.com/greek-stemmer/}} for Greek language competition, and then followed pre-processing steps as described above.
\subsection{Model Description}
We have used Ensemble model for all of the tasks mentioned above by combining CNN, self-attention, and LSTM based model.\par
\begin{algorithm}
\caption{Stacking Ensemble Algorithm}\label{alg:euclid}
\begin{algorithmic}[1]
\State \textbf{Input:} training data D= $\{x_{i},y_{i}\}^{m}_{i=1}$
\State \textbf{Output:} Ensemble classifier H
\State \textit{Step 1: Learn base level classifiers}
\State \textbf{for} \textit{t}=1 to \textit{T} \textbf{do}:
\State \quad learn \textit{$h_{t}$} based on dataset \textit{D}

\State \textbf{end for}
\State \textit{Step 2: Construct new dataset of predictions}
\State \textbf{for} \textit{i}=1 to \textit{m} \textbf{do}:
\State \quad \textit{$D_{h}$}= $\{x^\prime_{i},y_{i}\}$, where  $x^\prime_{i}=\{h_{1}(x_{i}),h_{2}(x_{i}),...,h_{T}(x_{i})\}$
\State \textbf{end for}
\State \textit{Step 3: Learn a meta classifier H}
\State learn H based on $D_{h}$
\State return \textbf{H}
\end{algorithmic}

\end{algorithm}
In the above algorithm, \textit{T} base level classifiers are trained on training dataset \textit{D}. These base classifiers are named as \textit{$h_{t}$}, where \textit{t} ranges from 1 to T. In step two new dataset \textit{$D_{h}$} is created for meta classifier where input is taken as base classifiers' output and model output as \textit{$y_{i}$}. Once the dataset is created meta classifier \say{H} is trained on dataset \textit{$D_{h}$}. In the end, algorithm returns trained meta-classifier \say{H}.\par 
\begin{figure}
        \centering
        \includegraphics[width=10cm,height=8cm]{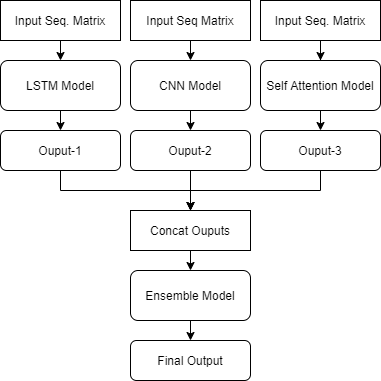}
        \caption{Ensemble Model}
        \label{fig:Model}
    \end{figure}
Stacking ensemble model is used for training RNN, CNN, Sequential self-attention with LSTM based architecture together in our model. In stacking, the algorithm takes output of sub-models as inputs and attempts to learn how to best combine inputs to get better output results. The idea of stacking is to learn several weak learners and combine them by training a meta-model to output predictions based on multiple predictions returned by these weak models \cite{Zhou:2012:EMF:2381019}. In the above algorithm we have three different Deep Learning models with labeled tweets separately. Then output of these models are used as independent variable for stacked model training and labels are same as previous steps.
\subsection{Parameters and Hyper-Parameters tuning of models}
\subsubsection{LSTM}
After preprocessing, dataset is split into two parts i.e. training set= 90\%, and validation set=10\%. For Recurrent Neural Network based model, we have used single LSTM layer with 256 cell units and then MaxPooling Layer to get maximum of all the tokens and then four dense layers with Dropout (dropout rate = 0.3) and BatchNormalization. Final layer is softmax or sigmoid layer depending upon the task. Softmax is used in Sentimix tasks and OffensEval English Subtask-C, and for rest of the subtasks sigloid is used. Model is trained on Focal loss function \cite{lin2017focal}, Adam optimizer \cite{kingma2014adam}, and metrics as accuracy and F1-score. Since maximum number of characters in a tweet are limited to 140 characters including space, we have taken the maximum number of words in a sentence to be 25 considering average lengths of words to be 5 and 3 to 4 spaces.
\subsubsection{CNN}
A stack of convolutional neural networks (CNN) is used for capturing the hierarchical hidden relations among embedding features. We trained data using CNN model with three convolution layer having filter sizes of (3,4, and 5) respectively, three max pooling layers with filter size of 2 and stride of 2, dense layers of size 4096 and 2048 with Dropout rate of 0.2. Dense layer is connected to softmax or sigmoid layer depending upon the task. Model is trained on Focal loss function \cite{lin2017focal}, Adam optimizer \cite{kingma2014adam}, and metrics as accuracy and F1-score. Tweets are padded in same way as in LSTM model.
 \subsubsection{Sequential self-attention model}
 We have used attention as explained in \cite{bahdanau2014neural}, after Gated Recurrent Unit layer \cite{chung2014empirical} by returning cell outputs from each steps. This model has 256 GRU cells and each cells return output state which are then fed to self-attention layer. Then there are same number of Dense layers and dropout rate as used in LSTM model. Rest parameters are same as in LSTM layer. 
\section{Experiments and Results}\label{experiments}
We performed our experiments on three Deep-Learning Models CNN, LSTM, and ensemble of CNN, LSTM, and self-attention. In Spanglish, we achieved F1-Macro of 0.770, precision of 0.749, and recall of 0.803 with Ensemble model on pre-processed data, whereas it was 0.709, 0.755 and 0.672 respectively  for raw text without pre-processing of the data. In Hinglish challenge, Ensemble model out performed other models on pre-processed dataset. Ensemble model on pre-processed data achieved F1-score of 0.682, precision of 0.695 and recall of 0.679 whereas same model when trained on raw data achieved 0.665, 0.681 and 0.665 respectively. From these results we can infer that cleaning steps involved helped to improve the results . We have also experimented our models with and without pretrained  Embeddings in Hinglish task but it did not help in improving the scores. We trained Hinglish embeddings by collecting code mixed Hinglish data from various sources such as blogs and scraping twitter data using Fasttext library \cite{bojanowski2016enriching}. Post evaluation on Spanglish task was not performed since gold labels were not released after the competition. Error analysis of Hinglish and Spanglish tasks is presented in \textbf{Appendix \ref{hinglish_error} and \ref{spanglish_error}} respectively. \par
Bert multilingual and Bert-uncased mode \cite{devlin2018bert} are trained and fine-tuned by adding three delta layers (dense) layers on top of pre-trained models. We trained Bert model using both ways one by freezing Bert pre-trained parameters and other by keeping parameters as trainable during the complete training process. From our experiments on Hinglish and Spanglish datasets using these models and techniques \footnote{\url{https://github.com/asking28/sentimix2020/blob/master/multilingual_bert.py}} we found that Bert-uncased model performed better than the Bert-multilingual model. Keeping the pre-trained parameters of Bert as trainable during the complete process performed better than freezing the Bert parameters during fine-tuning. We attribute this behavior of Bert to the difference in data distributions of Bert pre-training and Sentimix tasks. 
\begin{figure}[ht]

\begin{subfigure}{0.5\textwidth}
\includegraphics[width=1.1\linewidth, height=7cm]{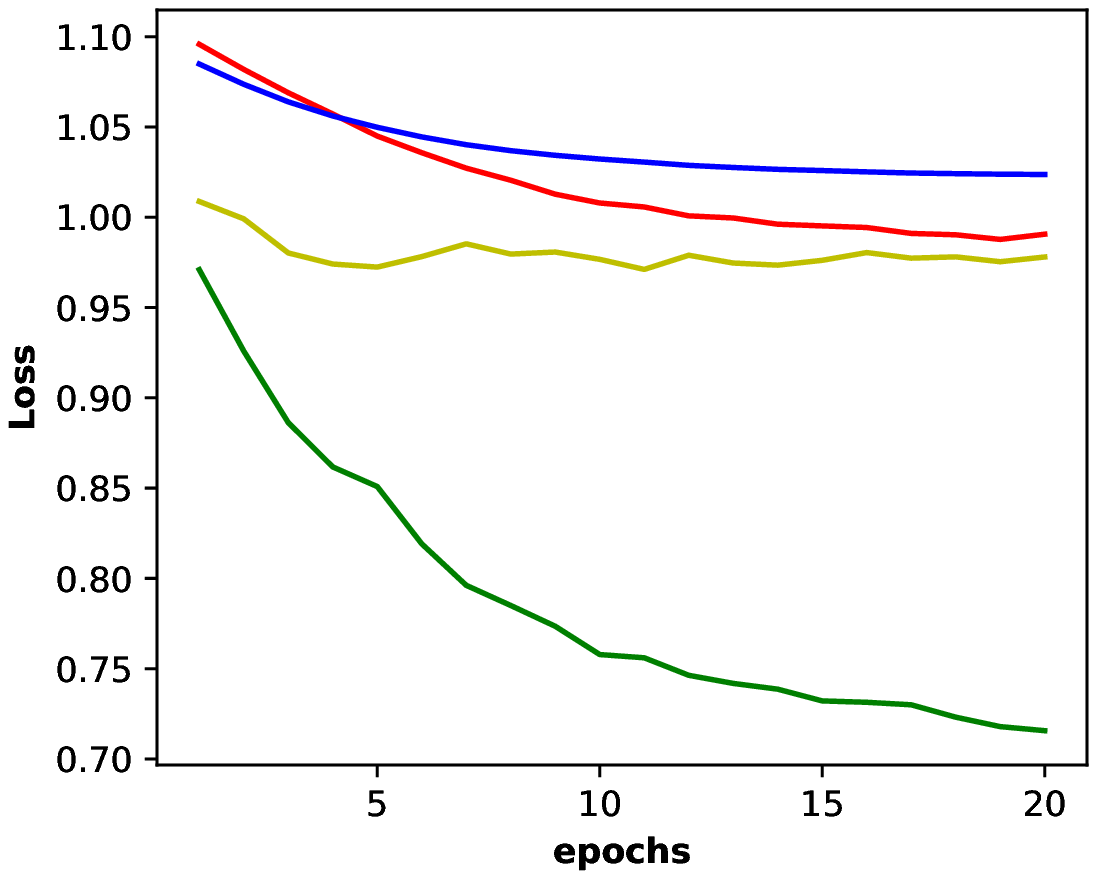} 
\caption{Loss}
\label{fig:loss1}
\end{subfigure}
\begin{subfigure}{0.5\textwidth}
\includegraphics[width=1.1\linewidth, height=7cm]{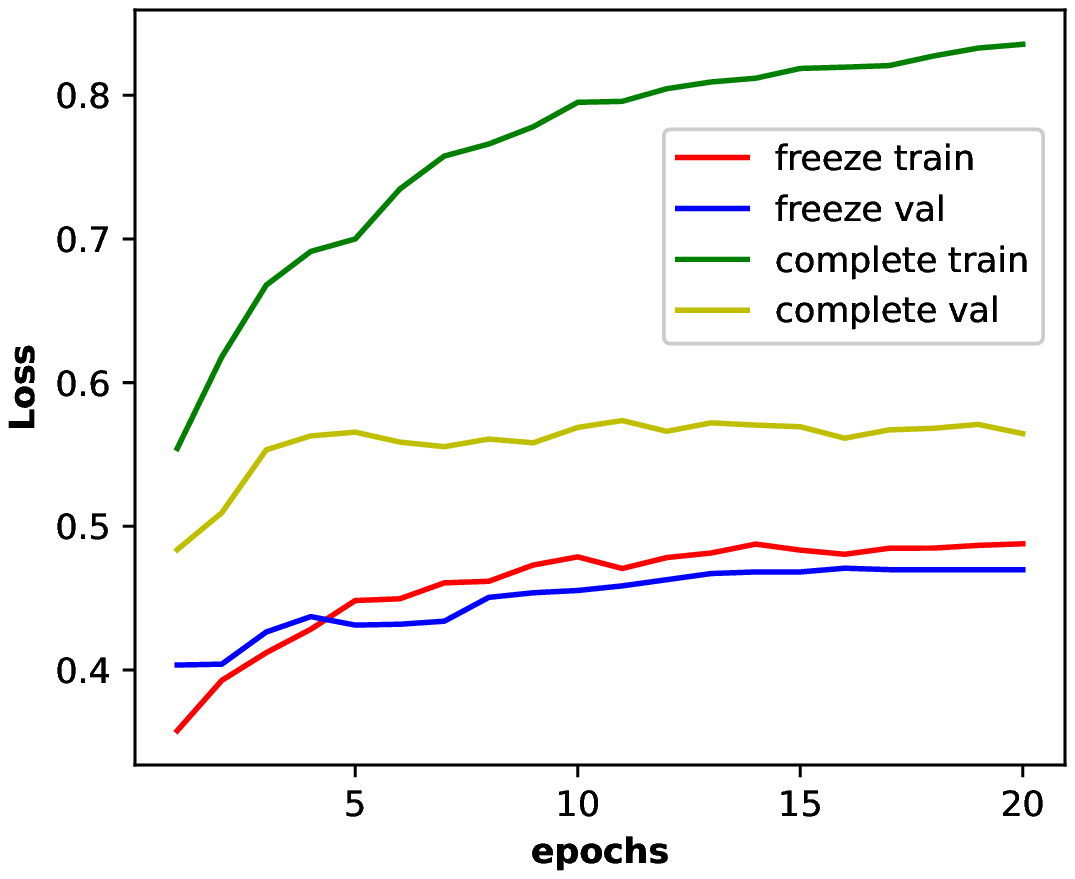}
\caption{Accuracy}
\label{fig:acc2}
\end{subfigure}

\caption{Bert Model Performance on Hinglish Dataset }
\label{fig:imagebert}
\end{figure}
\begin{figure}[ht]

\begin{subfigure}{0.5\textwidth}
\includegraphics[width=1.1\linewidth, height=7cm]{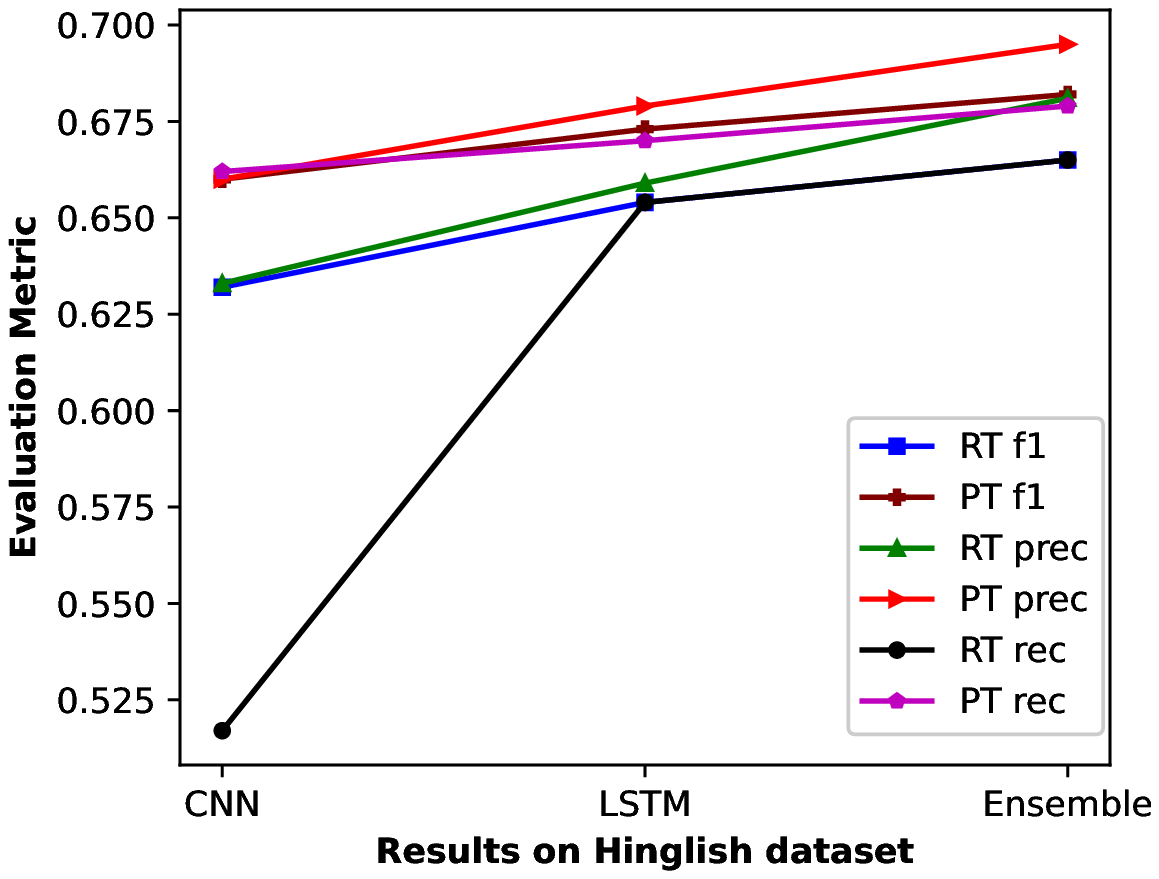} 
\label{fig:subim1}
\end{subfigure}
\begin{subfigure}{0.5\textwidth}
\includegraphics[width=1.1\linewidth, height=7cm]{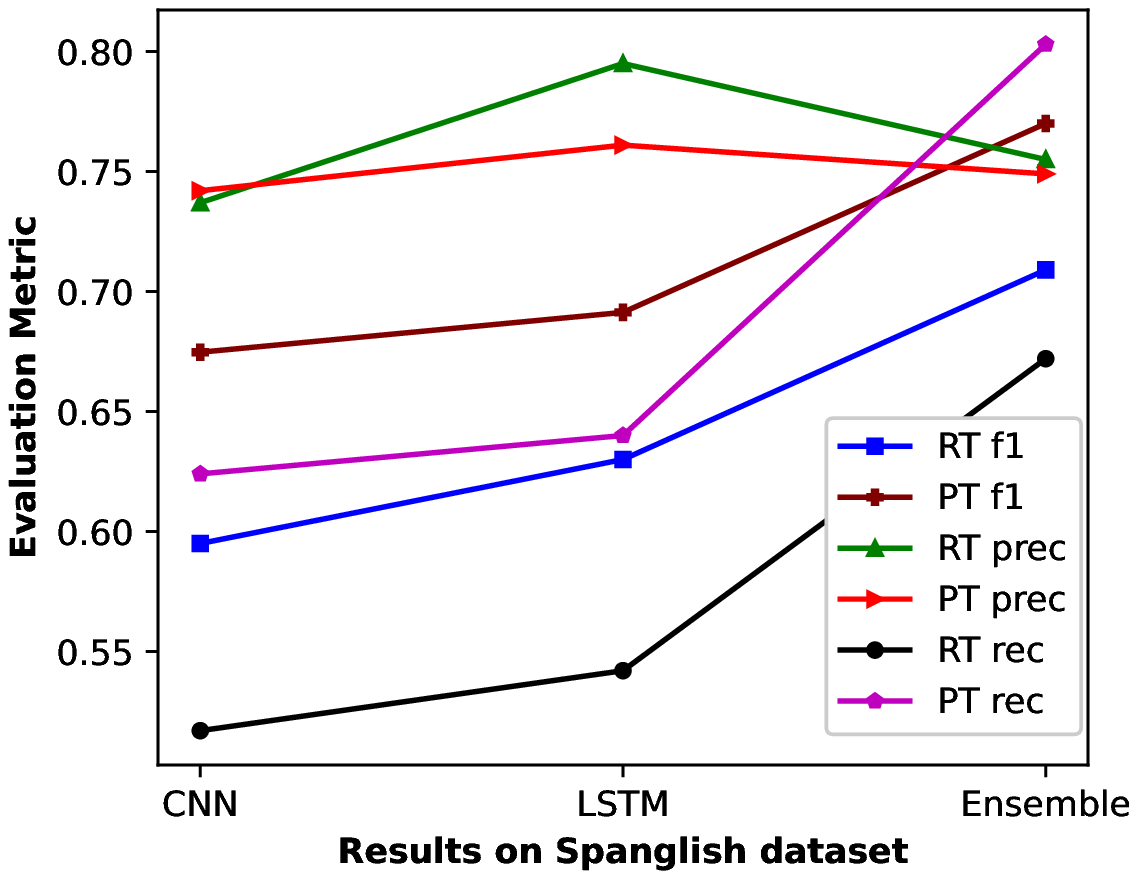}
\label{fig:subim2}
\end{subfigure}

\caption{Above plots show F1, Precision and recall when trained on LSTM, CNN and Ensemble models. Here f1 represents F1-macro, prec represents Precision, rec represents Recall, RT represents Raw Text, and PT represents Pre-processed Text. }
\label{fig:image2}
\end{figure}

\begin{table}[h]
\centering
\begin{tabular}{|c|c|c|c|c|c|c|}
\hline
\textbf{} &
  \textbf{\begin{tabular}[c]{@{}c@{}}Raw Text\\ (F1-Macro)\end{tabular}} &
  \textbf{\begin{tabular}[c]{@{}c@{}}Pre-processed\\ (F1-Macro)\end{tabular}} &
  \textbf{\begin{tabular}[c]{@{}c@{}}Raw Text\\ (Precision)\end{tabular}} &
  \textbf{\begin{tabular}[c]{@{}c@{}}Pre-Processed\\ (Precision)\end{tabular}} &
  \textbf{\begin{tabular}[c]{@{}c@{}}Raw Text\\ (Recall)\end{tabular}} &
  \textbf{\begin{tabular}[c]{@{}c@{}}Pre Processed\\ (Recall)\end{tabular}} \\ \hline
\textbf{CNN}      & 0.595 & 0.6747 & 0.737 & 0.742 & 0.517 & 0.624 \\ \hline
\textbf{LSTM}     & 0.630 & 0.6913 & 0.795 & 0.761 & 0.542 & 0.640 \\ \hline
\textbf{Ensemble} & 0.709 & 0.770  & 0.755 & 0.749 & 0.672 & 0.803 \\ \hline
\end{tabular}
\caption{Spanglish Testset Evaluation}
\label{tab:spanglist_testeval}
\end{table}
\begin{table}[h]
\centering
\begin{tabular}{|c|c|c|c|c|c|c|}
\hline
 &
  \textbf{\begin{tabular}[c]{@{}c@{}}Raw Text\\ (F1-Macro)\end{tabular}} &
  \textbf{\begin{tabular}[c]{@{}c@{}}Pre-processed\\ (F1-Macro)\end{tabular}} &
  \textbf{\begin{tabular}[c]{@{}c@{}}Raw Text\\ (Precision)\end{tabular}} &
  \textbf{\begin{tabular}[c]{@{}c@{}}Pre-Processed\\ (Precision)\end{tabular}} &
  \textbf{\begin{tabular}[c]{@{}c@{}}Raw Text\\ (Recall)\end{tabular}} &
  \textbf{\begin{tabular}[c]{@{}c@{}}Pre Processed\\ (Recall)\end{tabular}} \\ \hline
\textbf{CNN}      & 0.632 & 0.660 & 0.633 & 0.660 & 0.517 & 0.662 \\ \hline
\textbf{LSTM}     & 0.654 & 0.673 & 0.659 & 0.679 & 0.654 & 0.670 \\ \hline
\textbf{Ensemble} & 0.665 & 0.682 & 0.681 & 0.695 & 0.665 & 0.679 \\ \hline
\end{tabular}
\caption{Hinglish Testset Evaluation}
\label{tab:hinglish_testeval}
\end{table}

In Offenseval tasks, we performed post-evaluation experiments by tuning only few hyper-parameters in the model like changing class-weights and changing loss function. In Greek language subtask our model\footnote{\url{https://github.com/asking28/offenseval2020/blob/master/offens_2020_greek.ipynb}} achieved F1-score of 0.886, which is more than the highest F1-score of 0.852 achieved during evaluation period. In English Subtask-B, our model\footnote{\url{https://github.com/asking28/offenseval2020/blob/master/offens_task2_bilstm_attention.ipynb}} is able to achieve F1-score of 0.685 in post-evaluation, which was 0.580 in Evaluation period. In Subtask-A and C F1-score in post-evaluation period is 0.9084 and 0.5106 respectively, with very small difference from evaluation period. In Danish task, our model\footnote{\url{https://github.com/asking28/offenseval2020/blob/master/offens_2020_danish.ipynb}} was able to achieve F1-score of 0.6585 in post evaluation period and 0.613 during evaluation period. For Turkish and Arabic tasks there is small difference in the results in evaluation and post-evaluation experiments. Error analysis of English subtasks B and C are presented in \textbf{Appendix \ref{offens_b}, \ref{offens_c}} respectively.
\begin{table}[ht]
\centering
\begin{tabular}{|l|c|c|c|c|c|c|l|}
\hline
 &
  \textbf{\begin{tabular}[c]{@{}c@{}}English\\ Subtask-A\end{tabular}} &
  \textbf{\begin{tabular}[c]{@{}c@{}}English\\ Subtask-B\end{tabular}} &
  \textbf{\begin{tabular}[c]{@{}c@{}}English\\ Subtask-C\end{tabular}} &
  \textbf{Turkish} &
  \textbf{Arabic} &
  \textbf{Danish} &
  \textbf{Greek} \\ \hline
\textbf{Precision} & 0.8891 & 0.7028 & 0.5268 & 0.721 & 0.8014 & 0.643  & \textbf{0.882} \\ \hline
\textbf{Recall}    & 0.9437 & 0.6813 & 0.5013 & 0.707 & 0.8092 & 0.683  & \textbf{0.889} \\ \hline
\textbf{F1-Score}  & 0.9084 & 0.6855 & 0.5106 & 0.713 & 0.8052 & 0.6585 & \textbf{0.886} \\ \hline
\end{tabular}
\caption{Offenseval Testset  Post Evaluation}
\label{tab:offenseval_posttestevl}
\end{table}
\begin{figure}[H]
\includegraphics[width=0.7\textwidth, center]{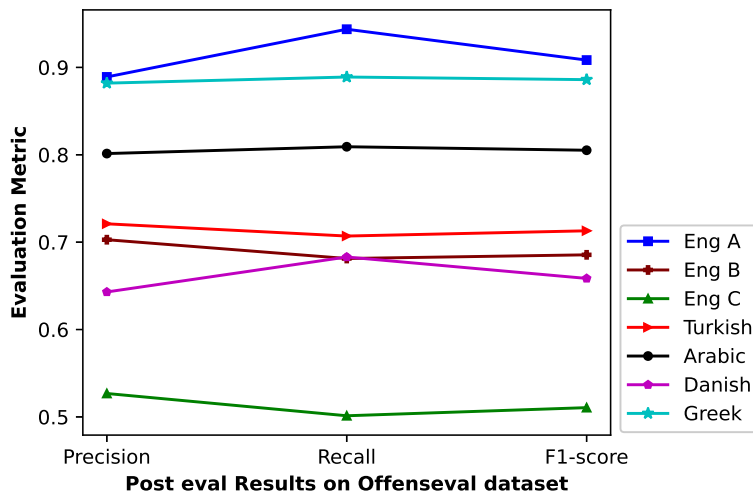}
\caption{This plot represents Precision, Recall and F1-score for all the OffensEval tasks. Eng A represents English Subtask-A and so on.}
\label{fig:figure2}
\end{figure}
\begin{table}[ht]
\centering
\begin{tabular}{|c|c|c|c|c|c|}
\hline
\textbf{\begin{tabular}[c]{@{}c@{}}English\\ Subtask-A\\ (F1-Macro)\end{tabular}} &
  \textbf{\begin{tabular}[c]{@{}c@{}}English\\ Subtask-B\\ (F1-Macro)\end{tabular}} &
  \textbf{\begin{tabular}[c]{@{}c@{}}English\\ Subtask-C\\ (F1-Macro)\end{tabular}} &
  \textbf{\begin{tabular}[c]{@{}c@{}}Turkish\\ (F1)\end{tabular}} &
  \textbf{\begin{tabular}[c]{@{}c@{}}Arabic\\ (F1)\end{tabular}} &
  \textbf{\begin{tabular}[c]{@{}c@{}}Danish\\ (F1)\end{tabular}} \\ \hline
0.905 &
  0.580 &
  0.514 &
  0.699 &
  0.800 &
  0.613 \\ \hline
\end{tabular}
\caption{Offenseval Testset Evaluation}
\label{tab:offenseval_testeval}
\end{table}

\section{Conclusion and Future work}\label{conclusion}
In this paper, we present description of the system that we have used in all OffensEval and Sentimix tasks. With our best model we were able to achieve third position in Spanglish task in evaluation period. In post evaluation experiments our model is able to achieve F1-score more than the highest score in Greek task evaluation period. From our experiments, we have found that pre-processing steps played a huge role in increasing F1-scores. Since we have used deep learning models, our model could not perform very well in the tasks where dataset was small like in English Subtask-C. In this work paper we present different data pre-processing  steps that played important role. In English Subtasks we experimented with pre-trained Embeddings \footnote{\url{https://github.com/FredericGodin/TwitterEmbeddings}} trained in twitter corpus, and found that pre-trained embeddings helped to increase F1-score in English sub-tasks but did not help in Hinglish task. The results obtained through our experiments in test data are lower than the results obtained in development set data. We inferred from our experiments that F1-score partly depends upon data distribution of different classes in training and development data which is used to tune hyper-parameters. In most of the tasks, data is not distributed equally among different classes. Exploratory data analysis reveals that there is huge difference in class distribution in the datasets.\par
Our system presents a solid baseline for Sentiment analysis of code-mixed languages and Offensiveness detection in multiple languages. In our future work, we plan to add handcrafted features along with current features and train it on different machine learning models. We also plan to explore techniques of data augmentation as Deep learning models need large amount of data to train. Corpus used to train code-mixed language models and languages other than English is very small as compared to corpus used to train English language models. Lot of research needs to be done in this direction.

\section{Acknowledgement}
 We deeply thank Ashutosh Singh, Gaurav Kumar, Himanshu Mangla, Suraj Tripathi, and Dr.Tribikram Pradhan for reviewing our work. These experiments have been performed on Google Colab. We thank Colab for providing GPUs and RAM free of cost.

\bibliographystyle{coling}
\bibliography{semeval2020}
\appendix
\appendixpage
\section{Hinglish Error Analysis}\label{hinglish_error}
This section presents the cases where the model did not give correct predictions and their possible reasoning. During the analysis of the Hinglish dataset on cross-validation data of 1869 tweets 40\% tweets were incorrectly predicted. Our study found that in most cases, either predicted class or ground truth class were labeled as neutral. In the hinglish dataset, our model failed to predict 747 tweets out of 1869 tweets in the validation dataset, and out of 747 tweets, 618 tweets (82\%) were labeled as \say{neutral} in either ground truth labels or predicted labels. From this, we can say that \say{neutral} class is a bit ambiguous in the Sentimix dataset. Consider a tweet- \say{rubika di umar mein aap se kaafi chota hun par am big fan of yours kabhi naseeb ne chaha to }. It is labelled as \say{neutral} whereas we feel that it should be labeled as \say{positive}. In most of the cases, our proposed model gets confused when it is/can be labeled as \say{neutral}.\par
\section{Spanglish Error analysis}\label{spanglish_error}

We found the same distribution in the Spanglish dataset. We performed validation in 2998 tweets, out of which 1025 were erroneous predictions. Out of 1025 tweets, 764 (74\%) of the tweets were labeled as \say{neutral} in either predicted or ground-truth class. Hence from our analysis, we found that labeling a tweet as \say{neutral} is somewhat ambiguous even for human annotators.\par
\section{OffensEval English Subtask-B Error Analysis}\label{offens_b}

We analyzed the categorization of offensive tweets on 37795 tweets as validation data. Out of this validation set, we got 7339 incorrect predictions, and out of these inaccurate predictions, 6912 tweets labeled as \say{targeted} were incorrectly predicted as \say{Un-Targeted}. Most of the targeted tweets have pronouns like \say{you, your, they, them, ur, u, she, her, these, him, his, he}, contain names of personalities who is targeted, and words like \say{people, bitch, boys, girls, variations of nigga }. Our model is biased towards such kind of words in a sentence. It can predict a sentence as \say{targeted}, which contains such type of terms. We feel that the dataset includes some incorrect labels, for example - \say{might fuck around and sleep without my feet covered}. It is not explicitly directed towards a person or a group. Our model sometimes fails to identify tweets directly targeting with names; for example, \say{This is some high level shit. Someone needs to dumb it down for Trump voters}. In most of the cases where our model fails to determine a tweet as targeted, the target is from \say{others} category where the objective is some event, situation, organization or an issue for example- \say{And here's another fucking breakdown}, \say{I'm sick of it all, April to August has been utter bullshit}.\par
\section{OffensEval English Subtask-C Error Analysis}\label{offens_c}
We analyzed target identification of OffensEval English subtask-C on 213 targeted tweets validation set. Out of 213 validation data points, 64 tweets' target is incorrectly predicted by our model. In this dataset we found that targets of some tweets are incorrectly labeled for example - \say{he should be ashamed of himself but he’s not because he’s \#zionel} is targeted towards an individual but labeled as other in the dataset and \say{\#arunjaitleystepdown he is most shameless \#fm in history of india and audacity and shamelessness with which is lies in public is disgrace to post.} is labeled as \say{Group} targeted but it is targeted towards an individual. Our model is able to classify these tweets as targeted towards an \say{Individual} correctly. Our model may be biased towards some pronouns, for example- \say{Dollar for a phone. you all are fucking dumb.} is classified as \say{Individual} targeted, but its correct label is \say{Group} targeted possibly due to the presence of \say{you} in the sentence. Also in tweet \say{anyway this game sucks}, model predicts as \say{Individual} targeted possibly because it is not able to decode what \say{this} refers to in the context, here \say{this} refers to an event \say{game}.
\end{document}